\documentclass[sigconf,authorversion,nonacm]{acmart}
\AtBeginDocument{%
  \providecommand\BibTeX{{%
    \normalfont B\kern-0.5em{\scshape i\kern-0.25em b}\kern-0.8em\TeX}}}

\copyrightyear{2023} 
\acmYear{2023} 
\setcopyright{acmlicensed}
\acmConference[KDD '23]{Proceedings of the 29th ACM SIGKDD Conference on Knowledge Discovery and Data Mining}{August 6--10, 2023}{Long Beach, CA, USA}
\acmBooktitle{Proceedings of the 29th ACM SIGKDD Conference on Knowledge Discovery and Data Mining (KDD '23), August 6--10, 2023, Long Beach, CA, USA}
\acmPrice{15.00}
\acmDOI{10.1145/3580305.3599843}
\acmISBN{979-8-4007-0103-0/23/08}

\usepackage{graphicx}
\usepackage{booktabs} 
\usepackage{subcaption}     
\usepackage{multirow}
\usepackage{amsfonts}       
\usepackage{nicefrac}       
\usepackage{microtype}      

\usepackage{soul}

\usepackage{etoolbox}

\newcommand{\pname}{IGB}

\usepackage{hyperref}

\begin{document}

\title{IGB: Addressing The Gaps In Labeling, Features, Heterogeneity, and Size of Public Graph Datasets for Deep Learning Research}

\author{Arpandeep Khatua}
\affiliation{%
  \institution{UIUC}
  \country{USA}
}

\author{Vikram Sharma Mailthody}
\affiliation{%
  \institution{NVIDIA}
  \country{USA}
}

\author{Bhagyashree Taleka}
\affiliation{%
  \institution{UIUC}
  \country{USA}
}

\author{Tengfei Ma}
\affiliation{%
  \institution{IBM Research}
  \country{USA}
}

\author{Xiang Song}
\affiliation{%
  \institution{AWS AI}
  \country{USA}
}

\author{Wen-mei Hwu}
\affiliation{%
  \institution{NVIDIA/UIUC}
  \country{USA}
}


\begin{abstract}
Graph neural networks (GNNs) have shown high potential for a variety of real-world, challenging applications, but one of the major obstacles in GNN research is the lack of large-scale flexible datasets. 
Most existing public datasets for GNNs are relatively small, which limits the ability of GNNs to generalize to unseen data. 
The few existing large-scale graph datasets provide very limited labeled data. 
This makes it difficult to determine if the GNN model's low accuracy for unseen data is inherently due to insufficient training data or if the model failed to generalize. 
Additionally, datasets used to train GNNs need to offer flexibility to enable a thorough study of the impact of various factors while training GNN models.

In this work, we introduce the \textbf{Illinois Graph Benchmark (IGB)\footnote{Accepted in KDD'23. DOI: 10.1145/3580305.3599843}}, a research dataset tool that the developers can use to train, scrutinize and systematically evaluate GNN models with high fidelity.
IGB includes both homogeneous and heterogeneous academic graphs of enormous sizes, with more than 40\% of their nodes labeled.  
Compared to the largest graph datasets publicly available, the IGB provides over 162$\times$ more labeled data for deep learning practitioners and developers to create and evaluate models with higher accuracy. 
The IGB dataset is a collection of academic graphs designed to be flexible, enabling the study of various GNN architectures, embedding generation techniques, and analyzing system performance issues for node classification tasks. 
IGB is open-sourced, supports DGL and PyG frameworks, and comes with releases of the raw text that we believe foster emerging language models and GNN research projects. 
An early public version of IGB is available at \url{https://github.com/IllinoisGraphBenchmark/IGB-Datasets}.

\end{abstract}




\received{2 February 2023}
\received[Accepted]{16 May 2023}

\maketitle
\section{Introduction}
Graph neural networks (GNNs) are a class of 
neural networks that operate on graph-structured data. 
GNNs have shown to be effective in addressing a variety of real-world applications such as fraud detection~\cite{amazonfraud, fraud-yelp, fraud}, recommendation systems~\cite{pinsage, etsy, linkedin}, predicting molecular and protein structure~\cite{drug1, drug2}, knowledge representation~\cite{knowledge}, and more recently helping in fine-tuning large language models~\cite{graphformer}.
Their popularity has led to the development of various optimized frameworks and libraries~\cite{dgl, distdgl, pyg} that enable the fast application of GNNs on new domains, making it easier for researchers and practitioners to leverage the power of GNNs in their work. 

However, high-quality frameworks and libraries are necessary but not sufficient for enabling fast research progress in GNN.
One of the major challenges in 
GNN research is the lack of large-scale datasets. 
This is because large graph datasets are typically propriety and most publicly available ones are rather small.
The small size of these datasets makes it difficult to train GNNs that can handle large-graph structures and prevents the use of powerful pre-training techniques~\cite{gat, gcn, bert,gnnsurvey}. 
These challenges make it difficult to fully leverage GNN potential and its applications. 

To address these challenges, recent work such as OGBN and MAG~\cite{ogblsc, ogbnmag} have proposed open large graph benchmark suites providing up to 244 million nodes and 1.7 billion edge graphs. 
These datasets contain a diverse set of graphs and have been widely used in the research community to benchmark GNNs' performance. 
However, most existing datasets, including OGBN, provide very limited labeled data. 
As GNN downstream tasks are often trained as supervised learning tasks, having large labeled data matters, especially for multi-label classification problems. 
However, both OGBN and MAG use Arxiv~\cite{ogblsc, ogbnmag} class labels which provide 1.4 million labeled nodes, meaning only about 1\% of the overall dataset is labeled!
With such small labeled data usage during training, it becomes challenging to determine if the GNN model's low accuracy for unseen data is inherently due to insufficient training data or if the model itself fails to generalize~\cite{low_label_1, low_label_data_2, low_label_data_3, low_label_data_4, low_label_data_5}.

Furthermore, the lack of flexible datasets hinders the researcher's ability to scrutinize and systematically evaluate the scalability of the GNN models, frameworks, and systems.
Ideally, a dataset should provide 
(a) capability to study the impact of embedding generation techniques and their properties on the GNN model's accuracy, 
(b) provide sub-datasets of varying graph sizes and embeddings maintaining consistent homophily, and 
(c) provide a range of multi-class classification tasks with varying degrees of complexity. 
Without such flexible datasets, it is difficult to train models on small graph datasets and then evaluate their accuracy and execution efficiency on larger data corpus, a common scenario in industrial settings. 
Moreover, the framework and system scalability problems encountered with small datasets are different from those 
with large datasets, making it challenging to study 
the system requirements of GNNs.

To this end, this work proposes \textbf{Illinois Graph Benchmark (\pname{})}, 
a research dataset tool that the researchers can use to scrutinize and systematically evaluate the GNN models and their execution efficiency. 
\pname{} provides access to enormous academic graph datasets for deep learning research consisting of both homogeneous and heterogeneous graphs, providing a diverse range of graph structures for training and evaluating GNN models. 
\pname{} homogeneous (\texttt{\pname{}-HOM}) graph dataset has up to 269 million nodes and about four billion edges. 
\pname{} heterogeneous (\texttt{\pname{}-HET}) graph dataset has up to 547 million nodes and about six billion edges. 
Both of these datasets come with more than 220 million labeled nodes created with a novel approach and with two complex node classification tasks (19 and 2983 unique classes). 
Compared to the largest graph dataset publicly available~\cite{ogblsc, ogbnmag}, \pname{} provides over 162$\times$ more labeled data, providing ample opportunities for GNN and deep learning practitioners to generalize the models to unseen data.

Furthermore, as a research tool, \pname{} offers flexible design choices for GNN model developer to investigate and systematically assess their models' performance and execution efficiency\footnote{Only node classification tasks are supported at the time of writing this paper}..
\pname{} provides variable-sized embeddings, can generate embeddings from different language models and provides a range of variable-size graphs with consistent homophily. 
To demonstrate the usefulness of such flexibility, we conduct an extensive ablation study to show the impact of node embedding generation on the GNN model accuracy. 
We find that using the Roberta NLP embeddings provides better accuracy on GNN models and that a larger embedding dimension further assists in the GNN model accuracy. 
However, for users who are memory constrained, we show that applying PCA dimensionality reduction can reduce the embedding vectors from 1024-dimensions to 384-dimensions, resulting in a 2.67$\times$ reduction in memory footprint with a maximum loss of 3.55\% in the GNN model accuracy.

Lastly, this work also discusses the system-level challenges faced when training and inferring GNN models on large graph datasets like \pname{}. As the dataset size increases, we observe a reduction in the effective GPU utilization due to the increased time spent waiting for the embedding sampling and gathering operation to complete. 
This is particularly pronounced with the full-scale IGB datasets which require over 1TB of memory space in a system and necessitates memory mapping from the storage. 
Our profiling shows the existing systems fail to adequately support efficient training and inference of GNN models when the datasets exceed host CPU memory. 

Overall this work makes the following key contributions:

\begin{enumerate}
    \item We propose \pname{}, a research dataset tool 
    that innovatively fills the critical gap in labeling, features, heterogeneity, and size of public graph datasets. 
    \item We show IGB offers flexible design choices enabling the exploration of different GNN designs, embedding generation schemes, the effect of labeled data, and how system performance evolves with increasing dataset size. 
\end{enumerate}

\pname{} is compatible with popular GNN frameworks like DGL~\cite{dgl} and PyG~\cite{pyg}, and comes with several predefined popular models. \pname{} is available for public usage including raw text used to generate embedding and with a public leaderboard soon~\cite{dummygithub}.

\section{Background and Motivation}
In this section, we provide a brief overview of GNNs and their applications. We then cover existing GNN datasets and discuss the importance of high-quality, large-scale datasets. 

\section{GNN overview}
\label{sec:gnnoverview}
GNNs are inspired by the idea of extending neural networks that are usually defined for vector inputs to graph-structured inputs. 
The key idea of GNNs is to pass messages between the nodes in a graph. 
They use a neural network to propagate information along the edges of the graph while updating the representations of the nodes as the model learns.
This enables the GNN to take into account the relationships between the nodes while learning their representations and properties.
GNNs can be used to solve a wide variety of tasks such as node classification, link prediction, and graph classification, commonly found in applications such as fraud detection~\cite{fraud, amazonfraud, fraud-yelp}, protein structure discovery~\cite{drug1, drug2}, and recommendation system~\cite{etsy, linkedin, pinsage}. 

GNNs can be applied to both homogeneous graphs and heterogeneous graphs. 
Different types of GNN models have been developed to operate on these graphs. 
For homogeneous graphs, the most popular GNN models are graph convolutional network (GCN)~\cite{gcn}, GraphSAGE~\cite{sage}, and graph attention network~\cite{gat}.
These models primarily differ in how they pass messages between the nodes to learn the representation. 
For heterogeneous graphs, the most popular ones are relational-GCN~\cite{rgcn} and relational-GAT(RGAT)~\cite{rgat}. 
While there are several GNN models, we can formalize all their update functions into a single equation: 
\begin{equation}
 h_v^{k} \gets \texttt{UPDATE}({ h_v^{k},  \texttt{SAMPLE}({h_{u}^{k-1}})})
\end{equation}

Depending on the number of types of relations and neighbor sampling method we can derive respective updated functions of the popular graph neural net models as shown in Table~\ref{tab:gnngeneraleq}, where,

\begin{tabular}{ll} 
\centering
${h_v}^k$ & $\gets$ embedding of node $v$ in the $k$ layer of the GNN. \\
$\sigma$  & $\gets$ non-linear activation. \\
${W_r}^k$ & $\gets$ weight matrix of GNN layer $k$ (for relation $r \in \mathcal{R}$). \\ 
          & \hspace{3ex}${W}^k$ for $r=1$.\\
$\mathcal{R}$ & $\gets$ set of all the relations in the graph. \\
$\alpha_{uv}$ & $\gets$ attention coefficient between node $u$ and $v$.\\
$c_{uv}$      & $\gets$ degree normalizing factor between node $u$ and $v$.\\
$\mathcal{N}^r(v)$ & $\gets$ neighbor of node $v$ (for $r \in \mathcal{R}$). $\mathcal{N}(v)$ for $r=1$.\\
\end{tabular}

Note that these equations ignore the self-node consideration and the bias term which remains the same for all models. 

\begin{table}
\centering
\caption{Popular GNN models' update function comparison.  }
\begin{tabular}{lr} 
\toprule
\texttt{GCN~\cite{gcn}} & $h_v^{k} \gets \sigma{\overbrace{\sum_{u \in \mathcal{N}(v)}}^\texttt{SAMPLE} \underbrace{ {\color{blue} (1/{c_{uv}})}\mathbf{W}^k}_{\texttt{UPDATE}} h_{u}^{k-1}}$ \\ 
\texttt{GraphSage~\cite{sage}} & $h_v^{k} \gets \sigma{ \underbrace{\mathbf{W}^k {\color{blue} AGGREGATE}}_\texttt{UPDATE}\overbrace{_{u \in { \color{blue} \mathcal{N'}(v)}} h_{u}^{k-1}}^\texttt{SAMPLE}}$ \\ 
\texttt{GAT~\cite{gat}} & $h_v^{k} \gets \sigma{\overbrace{\sum_{u \in \mathcal{N}(v)}}^\texttt{SAMPLE} \underbrace{ {\color{blue} \alpha_{uv}}\mathbf{W}^k}_{\texttt{UPDATE}} h_{u}^{k-1}}$ \\ 
\texttt{R-GCN~\cite{rgcn}} & $h_v^{k} \gets \sigma{\overbrace{{\color{blue} \sum_{r \in \mathcal{R}}} \sum_{u \in \mathcal{N}^{\color{blue} r}(v)}}^\texttt{SAMPLE} \underbrace{{\color{blue} (1/{c_{uv}})}\mathbf{W_{\color{blue} r}}^k }_\texttt{UPDATE} h_{u}^{k-1}}$ \\
\texttt{R-GAT~\cite{rgat}} & $h_v^{k} \gets \sigma{\overbrace{{\color{blue} \sum_{r \in \mathcal{R}}} \sum_{u \in \mathcal{N}^{\color{blue} r}(v)}}^\texttt{SAMPLE} \underbrace{{\color{blue} \alpha_{uv}}\mathbf{W_{\color{blue} r}}^k }_\texttt{UPDATE} h_{u}^{k-1}}$ \\
\bottomrule
\end{tabular}
\label{tab:gnngeneraleq}
\end{table}

\subsection{GNN Dataset Generation Techniques}
With the increasing popularity of GNNs, researchers have opted to generate GNN datasets based on their needs to simulate and test their model performance. 
Some of the popular GNN dataset generation techniques include: 
\textit{(a) Real-world graph datasets:} This technique takes an existing real-world database such as a transaction database or webpages and extracts graph structure from it~\cite{semanticscholar, ogblsc}.
\textit{(b) Graph sampling:} This technique selects a subset of nodes and edges from existing large graphs to create a sample dataset~\cite{suitesparse, amazonfraud}. 
\textit{(c) Synthetic graph generation:} This method creates synthetic graphs by randomly connecting nodes according to certain probability distributions such as power law or by using Kronecker graph generators~\cite{syngen, graphworld}. This technique can also be used as data augmentation to increase the dataset diversity.  
\pname{} datasets are created using real-world datasets and different configurations are created by performing graph sampling operations. We will describe the IGB dataset generation methodology in $\S$\ref{sec:design}.

\subsection{Existing Datasets And Their Problems}
\label{sec:motivation}

The existing dataset collection for GNNs are relatively small in size limiting the ability of GNN to generalize to unseen data~\cite{suitesparse, s2ag, s2orc, coragraph, bojchevski2017deep, citeseer, pubmed, mutag, co-datasets, aifb, ppi-reddit, ba-tree,fb15k-wn18, FB15k-237, BitcoinOTCDataset, qm7b, tu, gin, ged, entities-rellinkpred, aminer, wn18,  fbpagepage-github-wikipedia, emaileucore, amazonfraud}. 
More recently, work such as OGBN~\cite{ogblsc, snap, ogbnmag} have proposed datasets providing up to 244 million nodes and 1.7 billion edge graphs. 
These datasets contain a diverse set of graphs and have been widely used by the research community to benchmark GNN's performance. 

Table~\ref{tab:datasetcompare} summarizes the various types of publicly disclosed graph datasets.
We define flexibility in datasets as the ability to provide various configurations while maintaining homophily. 
This includes subgraphs with different node-degree distributions, variable-size embeddings, and different language models used to generate embeddings.  
Homophily refers to the tendency for individual nodes to be connected to other nodes with similar properties. 
With datasets that have similar homophily, researchers can better understand and develop GNN models by training with smaller graphs and evaluating accuracy on larger graphs. 

Table~\ref{tab:datasetcompare} covers both synthetic and real large graph datasets that are used to study the GNN model performance in both closed and open-source environments.
PinSAGE~\cite{pinsage} is one of the "few" publicly disclosed proprietary industrial datasets that have more than three billion nodes and eighteen billion edges with an unknown number of labels. 
Among the real-world open source datasets, MAG240M from OGB-LSC~\cite{ogblsc} provides the maximum number of nodes to evaluate multi-class node classification tasks. 
Yet, as noted from Table~\ref{tab:datasetcompare}, most existing datasets including OGBN datasets provide a tiny set of labeled data, provide no flexibility and the embeddings are generated in closed source.  

As the GNN downstream tasks are often trained as supervised learning tasks, having large labeled data helps in increasing the accuracy of multi-label classification problems. 
As several prior works have shown~\cite{low_label_1, low_label_data_2, low_label_data_3, low_label_data_4}, with such a small number of usable labeled nodes during training, it becomes challenging to determine if the GNN's low accuracy for unseen data is primarily due to insufficient training data or inability to generalize. 

\begin{table*}[htp]
\small
\centering
\caption{Comparison of IGB with the largest publicly disclosed existing dataset. *Labelled nodes as a percentage of total nodes. $input$ implies the sizes are dependent on the values provided to the generator.}
\begin{tabular}{lrcrrrrrccr} 
\toprule
Dataset & Date & Availability & Type  &\#Nodes & Labelled* & \#Edges & Dim  & RawText & Flexibility & Task\\ 
\midrule
\texttt{PinSAGE~\cite{pinsage}}                 & $2018$ & Private & Real &$3000$ M  & UNK.  & $18$ B  & 1024      & No & No & Multi-class classification \\
\texttt{papers100M~\cite{snap}}                 & $2020$ & Public   & Real &$111$ M   & $<1\%$ & $1.6$ B & 128       & No & No & $172$-class classification \\
\texttt{mag-240m~\cite{ogblsc}}                 & $2021$ & Public   & Real &$244$ M   & $<1\%$ & $1.7$ B & 768       & No & No & $153$-class classification \\
\texttt{GraphWorld~\cite{graphworld}}           & $2022$ & Public   & Syn  & $input$  & UNK.  & $input$  & N.A      & No & No & System Design \\
\texttt{SemanticScholar~\cite{semanticscholar}} & $2023$ & Public   & Real &$205$ M   & UNK.  & $~3$ B  & 768       & Yes & No & Multi-class classification \\
\texttt{SynGen~\cite{syngen}}                   & $2023$ & Private & Syn  & $input$  & N.A.     & $input$ & N.A       & No & Yes & System Design \\ \hline
\textbf{\texttt{IGB-HOM}}               & $2023$ & Public   & Real & $>260$ M & $>81\%$  & $~4$ B  & $128$ to  & Yes & Yes & $19$ or $2983$-class classification \\
\textbf{\texttt{IGB-HET}}             & $2023$ & Public   & Real & $>547$ M & $>40\%$  & $~6$ B  & $1024$  & Yes & Yes & and System Design \\
\bottomrule
\end{tabular}
\label{tab:datasetcompare}
\end{table*}

Besides, it is crucial to have access to flexible datasets when training GNN models in order to fully understand the impact of various factors on their accuracy. 
One key aspect is the generation of embeddings associated with the nodes or edges in the graph. 
As NLP models are used to generate the embeddings for GNN models, the quality of the NLP model can have a direct effect on the GNN model's accuracy and their downstream tasks. 
Thus, the datasets should provide mechanisms to study different embedding generation techniques along with methods to vary their proprieties such as node embedding dimensions, and pruning. 

Furthermore, as GNNs become increasingly popular, it is crucial to comprehend how their accuracy and efficiency (wall clock time) scale with the size and complexity of the graph. 
In this regard, researchers are developing optimized hardware and software frameworks tailored to the needs of the GNN applications~\cite{gnnsystem1, gnnsystem2, gnnsystem3, gnnsystem4, gnnsystem5}.
However, the lack of flexible datasets hinders their ability to evaluate the scalability of their systems as the datasets grow larger.  
Ideally, a dataset should provide sub-datasets of varying sizes that maintain the same graph structural properties to enable such comprehensive research on GNN systems.  
This is because the challenges faced with small datasets differ from those encountered with large datasets. 

For instance, training a small \pname{}-HOM graph can be completed within a reasonable time frame as the graph datasets and features can fit within a single system's memory.  
On the other hand, training a full version of the \pname{}-HET dataset with a single system with state-of-the-art software stack is currently challenging due to large memory requirements and complex software management techniques. 
To fully understand the potential and scalability of GNNs, it is essential to study their performance on a range of graph sizes and complexities.

\section{\pname{} Design}
\label{sec:design}
Illinois Graph Benchmark (\pname{}) datasets are designed to address the limitations discussed in $\S$~\ref{sec:motivation}. IGB datasets are created using real-world data extracted from Microsoft Academic Graph~\cite{mag} and SemanticScholar datasets~\cite{semanticscholar} as discussed in $\S$~\ref{sec:datasetgeneration}. 

\subsection{\pname{} Overview}
The \pname{} brings unique opportunities to assist in understanding the impact of dataset creation on the GNN model's accuracy. To this end, \pname{} addresses following set of challenges:
\label{sec:challengesgraphdataset}
\begin{enumerate}
    \item \textit{Open data licensing and text accesses}: 
    The use of datasets containing graphs, text, and embeddings to train emerging GNN models~\cite{dragon, rgcn, graphformer} is on the rise. 
    However, data sources used to create the datasets must have open licensing for ease of adaption and derivative product creation. \pname{} meets open-data-license requirements (ODC-By-1.0) and provides a collection of datasets containing graphs, text, and embeddings for GNN and language model training.  
    \item \textit{Large ground-truth:} \pname{} offers substantial ground-truth labels extracted from human-annotated data, eliminating the need for manually labeling millions of nodes. The majority of the \pname{} datasets are fully labeled, and the largest datasets have at least 40\% of their nodes labeled, which is achieved by fusing multiple databases to form a large labeled dataset while preserving the accuracy of the information.  

    \item \textit{Flexibility for ablation study:} The lack of flexible datasets limits the ability to evaluate the impact of the various components of the datasets on the performance of GNN models and understand their execution efficiency with different  hardware and software systems. \pname{} overcomes this by offering a flexible research tool providing variable-sized embeddings, the ability to generate embeddings from different language models, and a range of variable-sized graphs with consistent homophily. 

    \item \textit{Task complexity}: The \pname{} includes a range of multi-class classification tasks with varying degrees of complexity, which is crucial for evaluating the capabilities of GNN models. This is because while a GNN model can perform well on a 
    binary classification task, it might not be effective for more fine-grain classification tasks. This feature in \pname{} also enables the investigation of efficient transfer learning techniques for fine-tuning downstream tasks.
\end{enumerate}

\subsection{\pname{} Dataset Generation Methodology}
\label{sec:datasetgeneration}
Generating \pname{} datasets involves curating data from various sources. 
Each of the \pname{} dataset has a graph, ground truth labels, and node embeddings. 
We will first describe how we generate the graph by extracting information from real-world data. 

\textbf{Input Data:}
While there are a number of publicly available datasets that can be used as input source~\cite{pubmed, s2orc, s2ag, arxiv}, many of them are either small or lack the necessary information for building large-scale graph datasets for GNN application. 
To this end, \pname{} graphs are primarily created using the Microsoft Academic Graph (MAG)~\cite{mag} database and carefully added human-annotated labels from both MAG and SemanticScholar Corpus~\cite{semanticscholar} databases. Semantic Scholar data corpus is closely tied to MAG w.r.t to papers, as ~89.4\% of all Semantic Scholar paperIDs contain a field for unique reverse mapping with the MAG (a unique magPaperID).

The MAG database is particularly useful as it contains a wide range of relationships and information about different types of data points, including papers and authors, as well as information about paper citations, authorship, affiliations, and field of study. 
SemanticScholar corpus~\cite{semanticscholar}, while slightly smaller than MAG, also provides a similar set of functionalities. 
However, what particularly sets both of these datasets apart is the \textit{{explicit permission we received to release the raw text data under the ODC-By-1.0 licensing scheme}\footnote{It is important to note that the MAG dataset is now fully deprecated and we thank the MAG database owners for giving us explicit open-sourcing permission.}.}

The MAG comprises over 260 million entries for papers and about four billion relations representing a citation between two papers.  
The schema used by MAG has many tables among which paper, author, affiliation, URLs, conference/journals, and field of study tables are of interest. 
The paper table contains information such as title, published date, authors, citations, and names of journals or conferences where the article is published. 
The paper citations are stored in a file using coordinate (COO) graph storage format where the first paper ID cites the paired paper ID. 
The author table includes data on the author's name, affiliations, and papers they wrote. 
SemanticScholar corpus~\cite{semanticscholar} has a similar schema providing up to 205 million entries for papers and about three billion relations representing citation.

\begin{figure}[t]
    \centering
    \includegraphics[width=8.1cm]{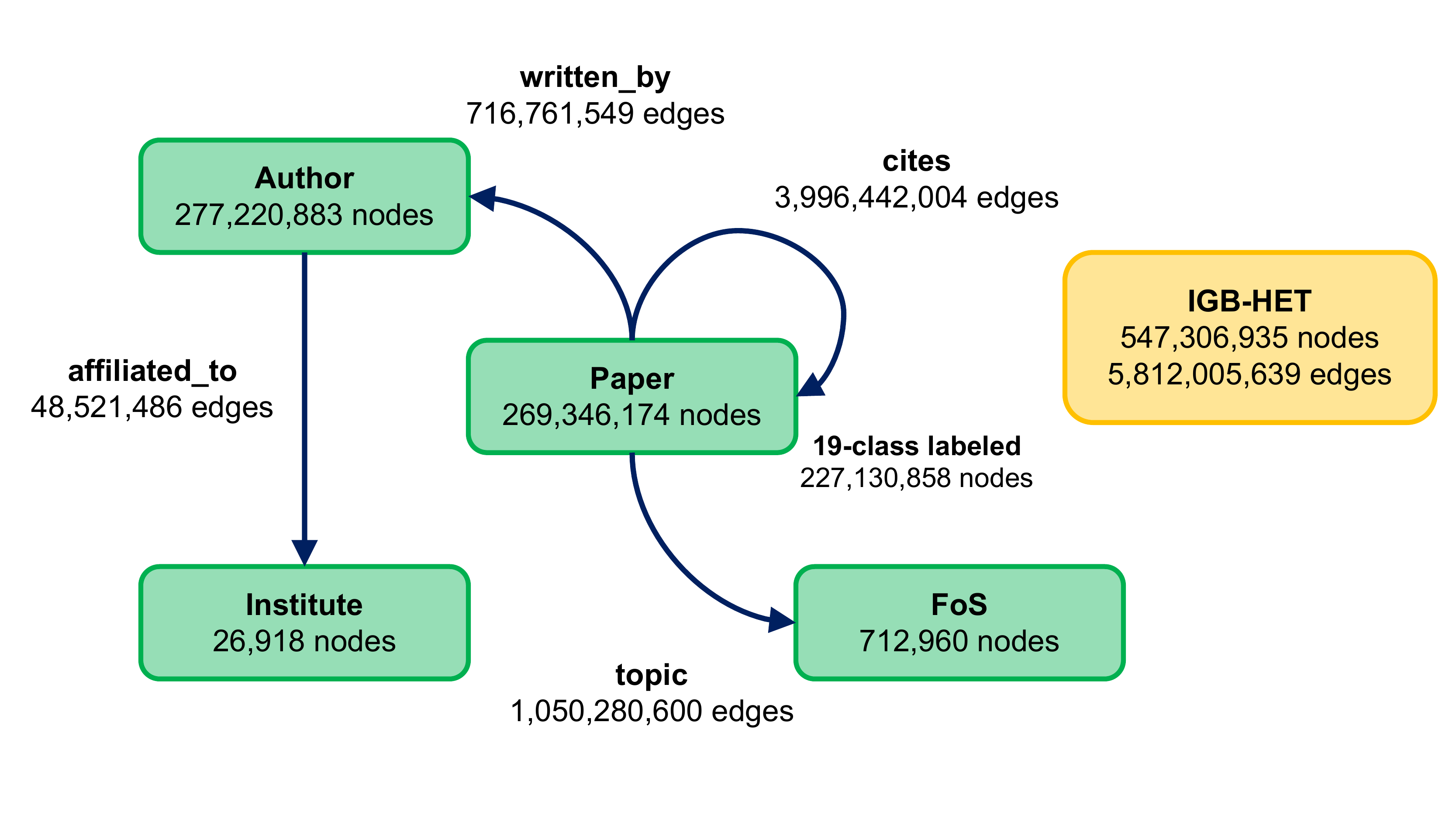}
    \vspace{-4ex}
    \caption{\texttt{IGB-HET} dataset schema.}
    \vspace{-4ex}
    \label{fig:hetero-schema}
\end{figure}

\textbf{IGB Dataset Graphs:} IGB provides two types of graphs: homogeneous and heterogeneous. The IGB homogeneous graph (\texttt{IGB-HOM}) is created by extracting only the paper nodes and the paper-cites-paper relation between these nodes. 
The IGB heterogeneous graph (\texttt{IGB-HET}) is created by extracting four different types of nodes: papers, authors, institutions, and fields of study. These different types of nodes are connected by various types of edges as shown in Figure~\ref{fig:hetero-schema}.

The IGB dataset collection (homogeneous and heterogeneous) graphs are created only using the MAG database. 
Although the Semantic Scholar and MAG databases exhibit a close association with paper nodes, the relationships with other node types such as authors and institutions poses a significant challenge. As a result only the relations from the MAG database are utilized during the construction of IGB graphs. 
Combining the graph structure using Semantic Scholar and MAG would give us an additional 20 million (~0.5\%) edges but will introduce data inconsistencies. 
For example, MAG and SemanticScholar have author names mismatches despite referring to the same individual. 
Some of these inconsistencies can be addressed by emerging entity resolution techniques which are still being discussed in the research. 
We hope by releasing the raw data as a part of IGB we can help build better entity resolution techniques and eventually assist in building much larger graphs.

\begin{table*}[t]
\small
\centering
\caption{{\texttt{IGB-HOM}} dataset collection metrics. Maximum memory sizes are reported (embd.).}
\vspace{-3ex}
\begin{tabular}{lrrrrrrrrr} 
\toprule
{\texttt{IGB-HOM} (sizes)} & \#Nodes & \% Labeled & \#Edges & Degree & Homophily & Emb-dim & \#Classes  & Mem.(graph/embd./total) \\ 
\midrule
\texttt{tiny}   & $100,000$     & $100$  & $547,416$        & $234/1/5.47$      & $56.79\%$ & $128-1024$ & $19/2983$ & $6.9$ MB/$393$ MB/$400$ MB  \\
\texttt{small}  & $1,000,000$   & $100$  & $12,070,502$     & $4,292/1/12.07$   & $47.75\%$ & $128-1024$ & $19/2983$ & $185$ MB/$4.0$ GB/$4.1$ GB  \\
\texttt{medium} & $10,000,000$  & $100$  & $120,077,694$    & $22,315/1/12.00$  & $59.93\%$ & $128-1024$ & $19/2983$ & $1.8$ GB/$39.0$ GB/$40.8$ GB\\
\texttt{large}  & $100,000,000$ & $100$  & $1,223,571,364$  & $73,248/1/12.20$  & $58.27\%$ & $128-1024$ & $19/2983$ & $19$ GB/$400$ GB/$401.8$ GB \\
\texttt{IGB-HOM}   & $269,346,174$ & $84.3$ & $3,995,777,033$  & $277,194/1/14.90$ & $51.79\%$ & $128-1024$ & $19/2983$ & $56$ GB/$1.1$ TB/$1.15$ TB  \\
\bottomrule
\end{tabular}
\label{tab:igbhomogeneous}
\end{table*}
 
\begin{table*}[t]
\small
\centering
\caption{{\texttt{IGB-HET}} dataset collection metrics. Reported maximum memory sizes with 1024-dim embeddings (embd.).}
\vspace{-3ex}
\begin{tabular}{lrrrrrrrr} 
\toprule
{\texttt{IGB-HET} (sizes)} & \#Total Nodes & \#Paper Nodes & \#Author Nodes & \#Inst. Nodes & \#FoS Nodes & \#Total Edges & \#Classes & Mem.(graph/embd./total) \\ 
\midrule
\texttt{tiny}  & $549,999$     & $100,000$     & $357,041$     & $8,738$  & $84,220$  & $2,062,714$     & $19/2983$ & $31.5$ MB/$2.19$ GB/$2.2$ GB\\
\texttt{small} & $3,131,266$   & $1,000,000$   & $1,926,066$   & $14,751$ & $190,449$ & $26,488,616$    & $19/2983$ & $391$ MB/$12.16$ GB/$13$ GB \\
\texttt{medium}& $25,982,964$  & $10,000,000$  & $15,544,654$  & $23,256$ & $415,054$ & $249,492,193$   & $19/2983$ & $3.8$ GB/$100.8$ GB/$104$ GB\\
\texttt{large} & $217,636,127$ & $100,000,000$ & $116,959,896$ & $26,524$ & $649,707$ & $2,104,750,425$ & $19/2983$ & $30.8$ GB/$844$ GB/$874$ GB \\
\texttt{IGB-HET}  & $547,306,935$ & $269,346,174$ & $277,220,883$ & $26,918$ & $712,960$ & $5,812,005,639$ & $19/2983$ & $85$ GB/$2.2$ TB/$2.28$ TB  \\
\bottomrule
\end{tabular}
\label{tab:igbheterogenos}
\end{table*}

\textbf{IGB Ground-Truth Labels:}
The challenge of obtaining a large collection of real-world ground truth labels is a major issue pertaining to dataset generation especially when manual annotation at the million scales is unattainable. 
Currently, the largest graph datasets such as MAG240M~\cite{ogblsc} and OGBN-paper100M utilize ArXiv~\cite{arxiv} labels and have a maximum of 1.4 million nodes categorized into 172 paper topics and eight subject areas. 
However, such a tiny set of labeled nodes is insufficient for real-world applications as the model cannot accurately predict unseen data. 

To address this, IGB extracts human annotated label information from the MAG~\cite{mag} and SemanticScholar~\cite{semanticscholar} databases. 
Extraction of information and aligning two databases pose a challenging problem. 
First, the two databases are distinct, and the nodes from both sides must be correctly aligned.
We address this by leveraging the reverse mapping of MAG paper IDs from the SemanticScholar database and merging the databases to create the IGB graph.

Second, the two databases have different labeling methodologies. 
To handle this issue, we carefully create a union of distinct labels from the MAG and SemanticScholar databases. 
Third, the two databases can have different distinct labels for each paper and can create a merge conflict. 
We address this merge conflict by manually checking the very small number of nodes that have this issue and assigning the right label that appropriately defines the paper from the union of distinct labels from MAG and SemanticScholar databases. 
Using this merged database, IGB datasets can cover a large portion of the data with labeled nodes. 
As shown in Table~\ref{tab:datasetcompare}, IGB provides more than 81\% and 40\% of nodes labeled for homogeneous and heterogeneous graphs. 

The paper nodes in the IGB dataset collection are labeled using the below labeling rules. These rules are carefully crafted to ensure label consistency in the data while merging labels from both MAG and Semantic Scholar databases: 

\begin{itemize}
    \item \textit{MAG has labels for the given paper node (~$58\%$ of papers):} Regardless of whether Semantic Scholar has a label or not, if MAG has a label, we use it directly.
    \item \textit{MAG does not have a label and Semantic Scholar has a label (~$25\%$ of papers):} If MAG does not provide a label but Semantic Scholar has a label and a unique magID reverse mapping, we use the label from Semantic Scholar.
    \item \textit{Neither MAG nor Semantic Scholar provides a label (~$17\%$ of papers):} For papers that do not have labels in either MAG or Semantic Scholar, we consider them unlabeled.
\end{itemize}
As we prioritize MAG labels over Semantic Scholar labels, we do not have label consistency issues in terms of under/over merges.

\textbf{IGB Downstream Tasks:}
The IGB dataset collections are designed for multi-class classification problems with two different numbers of classes (19 and 2983) depending on the degree of complexity. 
The $19$ class task is curated by combining classes from MAG and SemanticScholar and mapping them into a common structure. 
Examples of 19-class labels are history, geology, economics, and many more. 
The $2983$ class task is created by bucketing all papers with the same set of paper topics from the set of labels provided by SemanticScholar corpus. Examples of class labels are pediatrics, criminology, and computer engineering.
The 19-class task is intended for model development and testing while the 2983-class task can be used to stress test the models and develop robust GNN models for noisy real-world data. 
Although the IGB dataset comprises node classification problems to solve, it is easy to extend the downstream task to train for edge prediction tasks such as citation recommendation or reviewer recommendation. 
For instance, an edge prediction task can be constructed by masking our existing edges in the graph, training an edge prediction model and use the masked edges as the ground truth output. 

\textbf{IGB Embedding Generation: }
GNN models operate on graph structure and their embeddings. 
Embeddings can be associated with a node or an edge and capture the nature of the nodes as well as the nature of relationship between the nodes and help the GNN modes to extract deeper semantic information from the graph.  
In the past, one-hot vectors and word dictionaries were used to generate these embeddings. 
However, with the introduction of word2vec embeddings and more recently deep learning-based word and text embeddings, GNNs are being initialized with embeddings generated using foundational language models. 

For IGB datasets, we generate embeddings for each node in the graph
\footnote{Although IGB can generate edge embeddings, from the data management perspective it is quite challenging due to the sheer size of the generated dataset.}.
Node embeddings are generated by passing the paper titles and abstracts through a Sentence-BERT model~\cite{sent_bert} and obtaining a 1024-embedding vector for each paper node. 
Sentence-BERT, based on Siamese network, can generate semantically meaningful sentence embeddings by modifying a pre-trained BERT model~\cite{bert} such as RoBERTa~\cite{roberta} or GPT~\cite{gpt2} while maintaining high accuracy with least run-time overheads. 

As IGB graphs are extracted from scientific data~\cite{mag, semanticscholar}, it is possible to create domain-specific embeddings instead of general language model-based embeddings. 
For scientific text, SPECTER~\cite{specter}, an embedding model created using SciBERT~\cite{SciBERT}, a variant of BERT, can be used. 
SPECTER~\cite{specter} uses positive and negative samples of the papers from the SemanticScholar corpus~\cite{semanticscholar} to optimize the embeddings for the scientific text, leading to a 1.5\% gain in the F1 score in the 19-class task compared to the Sentence-BERT model when tested on papers from the MAG dataset.

We chose to use sentence-BERT embeddings trained on web data as the default IGB embeddings, as we want to benchmark the GNN model's ability to extract structural information from the embeddings that do not contain inherent fine-tuning towards a specific domain. 
This also reflects the practical considerations of real-world industrial settings where fine-tuning language models for each domain-specific task is time-consuming and expensive. 

The IGB author node embeddings are generated by taking the average of the node embeddings of all the papers written by the author, a methodology followed 
by Hu, et al.~\cite{ogblsc}. 
Institute node embeddings are generated similarly taking the average of the node embeddings of all the authors affiliated with the particular institute. 
Field of study node embeddings is generated using the topic name provided in the field of study in the database.

\subsection{IGB Datasets}
\label{sec:igbdatasets}
The IGB dataset suite offers five datasets for both homogeneous and heterogeneous graphs with varying sizes for deep learning practitioners as shown in Table~\ref{tab:igbhomogeneous} and Table~\ref{tab:igbheterogenos}. 
Each dataset is larger than the previous one (by about an order of magnitude) and is designed for a specific purpose.
Each of the smaller datasets are created by carefully sampling the graph such that we maintain similar homophily across different dataset variations. 
Homophily of the \pname{} dataset varies from 47.75\% to 59.93\% consistent with the homophily of prior released citation graphs~\cite{ogblsc, arxiv}.
The tiny dataset is meant for testing and development of GNN models and can be run on a laptop or mobile or an edge device. 
The small and medium datasets are ideal for training and testing new GNN models during development and can be trained with lower to mid-end GPUs or a powerful CPU. 
The small and medium datasets are also representative of large-scale labeled datasets currently available to the public~\cite{ogblsc}. 
The large dataset requires significant computing resources and can be trained on high-end accelerators and is suitable for building robust GNN models and testing by system designers. 
The full dataset is a massive dataset for GNN developers to construct practical models and stress test the distributed training platforms.  

Each dataset is initialized with 1024-dimensional node embeddings and has two different output classes that increase the difficulty to stress testing the GNN models and optimizing model development. 
The datasets are randomly split with 60\% for training, and 20\% each for validation and testing.
The dataset also includes the year of publication metadata for every paper node in case the user wants to set a specific splitting rule. 

\section{\pname{} Dataset Case Studies}
\label{sec:result} 
The \pname{} dataset flexibility enables extensive ablation study to understand the impact of dataset generation technique on the GNN model performance. 
Through these case studies, we make the following key observations:
 
\begin{itemize}
    \item GNN models for node classification tasks observe up to 12.96\% boost in their accuracy as the fraction of labeled nodes in the dataset increased from 1/150 to fully labeled.
    \item Using an NLP model for initializing node embeddings results in 40\% increase in GNN accuracy over random node embeddings.
    \item Among the different generic NLP model tests, RoBERTa NLP model provides the best overall performance. 
    \item Reducing the embedding dimension from 1024-dim to 384-dim results in 2.67$\times$ memory saving with a maximum 3.55\% average loss in the GNN models accuracy.
\end{itemize}

\subsection{Setup}
\label{sec:syssetup}
\textbf{Models:} We present performance benchmarks for three commonly used GNN models (GCN~\cite{gcn}, GraphSAGE~\cite{sage}, and GAT~\cite{gat}) on the IGB homogeneous dataset and for three popular GNN models (RGCN~\cite{rgcn}, RSAGE\footnote{RSAGE, a GraphSAGE extension to relational graphs is reported for the first time.} and RGAT~\cite{rgat}) on the IGB heterogeneous dataset. 
All models are trained with $0.01$ learning rate with two layers. Most of the models are trained to three epochs and numbers are reported. 
Unless explicitly stated, we used a batch size of 10K to maximize GPU utilization and used IGB-medium as the default dataset. 
Lastly, most evaluations are reported with homogeneous IGB datasets but are equally applicable to heterogeneous datasets.

\textbf{System:} We conducted our evaluations on a high-performance server system with dual AMD EPYC 7R32 processors, 256GB of DRAM, and an NVIDIA A100-40GB PCIe GPU with NVMe SSDs. The experiments were carried out using the DGLv0.9 framework~\cite{dgl} built on top of PyTorch~\cite{pytorch} as obtained from the NVIDIA NGC repository. \textit{To reflect real-world cost considerations in industry, all experiments were run on a single EC2 instance.}

\subsection{Impact Of Labeled Nodes}
\begin{figure}[t]
    \centering
    \includegraphics[width=8cm]{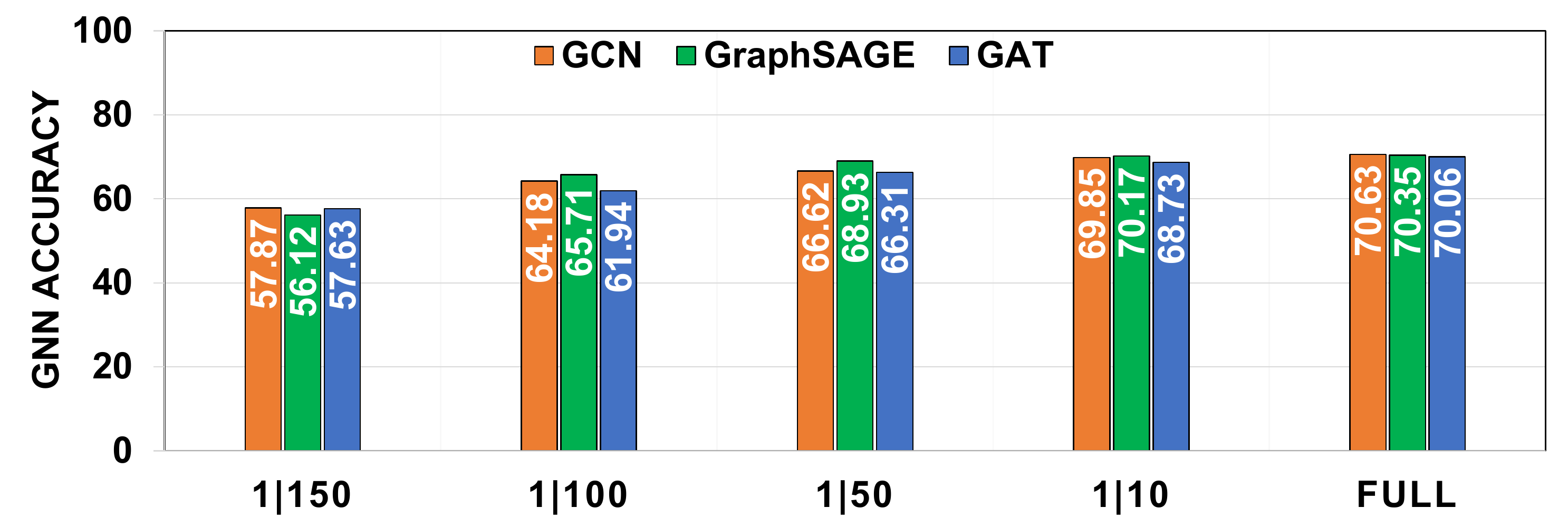}
    \vspace{-1ex}
    \caption{{Impact of labeled nodes on GNN model performance using \texttt{IGB-HOM-medium} dataset. Having more labeled nodes during training improves the model's accuracy. Full refers to all nodes labelled in \texttt{IGB-HOM-medium}.}}
    \label{fig:labelvariation}
\end{figure}

\begin{table}[t]
\centering
\small
\caption{Impact of labeled nodes on IGB variants. Average accuracy across (R)GCN, (R)SAGE and (R)GAT models are reported for the 1/150 and full labeled (lbl) dataset. }
\label{table:fractiondataset}
\begin{tabular}{lrrr|r} 
\toprule
Dataset & \# Class & Full-lbl acc & ${1}/{150}$ lbl acc & Diff. \\ 
\midrule
\texttt{IGB-HOM-medium} & $19$ & $70.34\%$ & $58.15\%$ & $12.18$ \\
\texttt{IGB-HOM-medium} & $2983$ & $62.35\%$ & $49.39\%$ & $12.96$ \\
\texttt{IGB-HET-medium} & $19$ & $72.35\%$ & $61.66\%$ & $10.69$ \\
\texttt{IGB-HET-medium} & $2983$ & $72.46\%$ & $63.26\%$ & $9.20$ \\
\hline
\label{tab:labelvariation}
\end{tabular}
\end{table}

In Section~\ref{sec:motivation}, we emphasized the importance of labeled data. 
In this section, the impact of labeled data is evaluated for the IGB medium dataset. 
To achieve this, sub-datasets with varying fractions (i.e., 1/150, 1/100, and so on.) of labeled data were created 
and then trained and evaluated using the GCN, GraphSAGE and GAT models on the 19-class node classification task. 
$1/150$ labeled sub-dataset is representative of the largest publicly available labeled dataset, MAG240M~\cite{ogblsc} and is created by picking one labelled node every 150 nodes.
As shown in Figure~\ref{fig:labelvariation}, the model accuracy improved by up to 10\% as more labeled data was added.  
Increasing the classification task complexity from 19-class to 2983-class shows a similar trend in the performance.
This is expected behavior as these models are trained as supervised learning tasks and more labeled data should help in boosting the model's performance. 

The experiment was also performed on three sub-variants of the homogeneous dataset and on medium size IGB heterogeneous dataset. 
As shown in Table~\ref{tab:labelvariation}, adding more labeled data helps improve the model performance significantly even with the IGB dataset variants including heterogeneous datasets.
This is promising mainly because, with datasets that are 100\% labeled, the GNN developer can develop more accurate models and not be constrained by a lack of labeled data.

\begin{table}[t]
\centering
\small
\caption{Impact of labeled nodes on IGB using SAGE compared to MLP.}
\label{table:fractiondataset}
\begin{tabular}{lrr|r} 
\toprule
Dataset & MLP (2 layer) & SAGE & Diff. \\ 
\midrule
\texttt{IGB-HOM-tiny} (1/150 labeled) & $47.89\%$ & $53.97\%$ & $6.08$ \\
\texttt{IGB-HOM-tiny} (all labeled) & $66.00\%$ & $69.38\%$ & $3.38$\\
\texttt{IGB-HOM-small} (1/150 labeled) & $52.90\%$ & $57.20\%$ & $4.30$  \\
\texttt{IGB-HOM-small} (all labeled) & $73.18\%$ & $75.49\%$ & $2.31$ \\
\texttt{IGB-HOM-medium} (1/150 labeled) & $52.11\%$ & $58.15\%$ & $6.01$ \\
\texttt{IGB-HOM-medium} (all labeled) & $70.75\%$ & $70.30\%$ & $-0.45$ \\
\hline
\label{tab:mlpvsage}
\end{tabular}
\end{table}

\textbf{Comparison with MLP: } The strength of GNNs, typically manifests when the label fraction is relatively small. According to the table~\ref{tab:mlpvsage} above, GNN models exhibit better performance compared to the supervised MLP baseline for the homogeneous graph, when the labeled data constitutes a small fraction of the overall dataset. However, when the dataset is fully labeled, the benefit provided by the GNN models is comparable to the naive MLP baseline with two layers.

While the accuracy of the GNN model surpasses that of the MLP baseline when a small fraction of the dataset is labeled, its throughput is an order of magnitude lower than that of the MLP baseline. From a system design perspective, this raises the question of whether we should consider developing a more complex MLP architecture for homogeneous graph datasets. 

\subsection{Embedding Generation Ablation Study}
Embeddings are key to the GNN model's performance but none of the prior work has shown how they impact GNN's performance.  
In this section, we will discuss the importance of embedding, and its generation process in-depth. 
\begin{figure}[t]
    \centering
    \vspace{-4ex}
    \includegraphics[width=8cm]{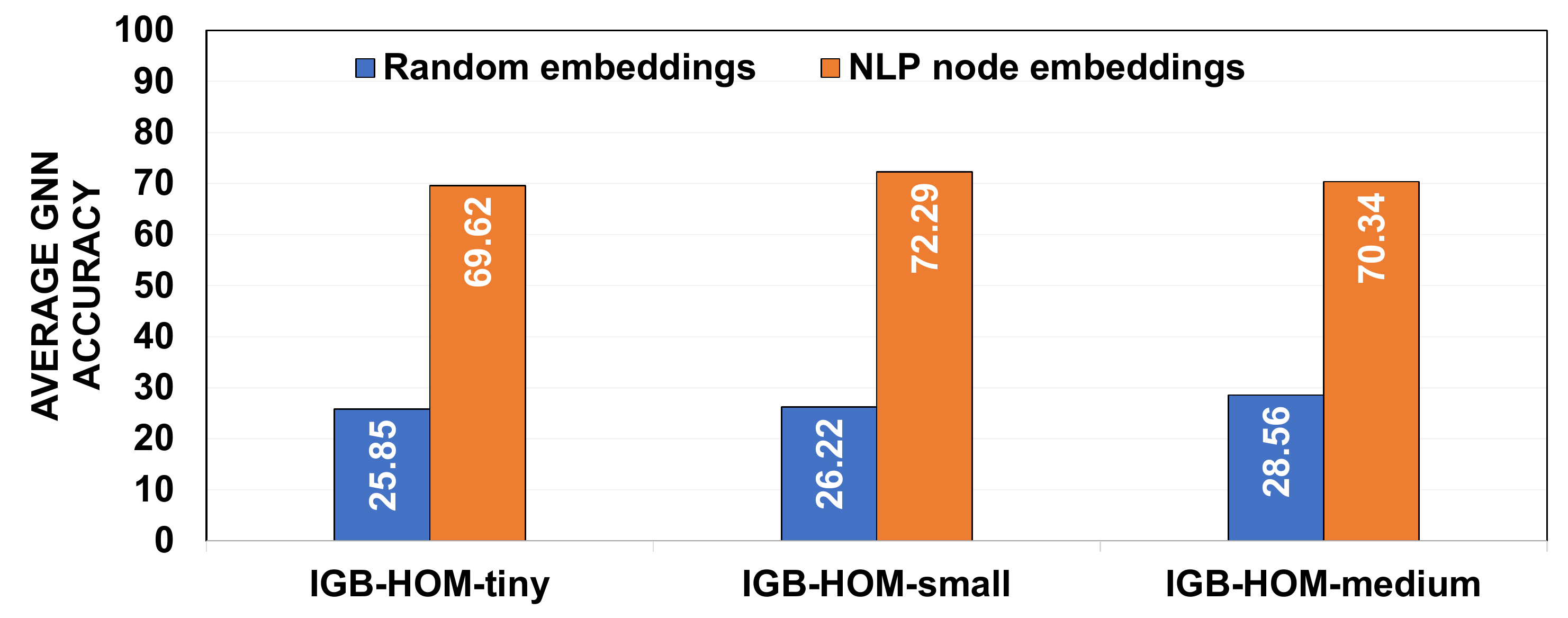}
    \vspace{-1ex}
\caption{NLP embeddings help the GNN model to learn both structural and node properties while random embeddings can only learn from the structure.}
\label{fig:randominit}
\vspace{-4ex}
\end{figure}

\textbf{Random vs. NLP embeddings}
The GNN performance depends on the node information provided to the model, which is initialized as node embeddings. 
In this evaluation, we compared the difference in performance between randomly initialized vectors and RoBERTa~\cite{roberta} NLP-based embeddings on three IGB homogeneous datasets as shown in Figure~\ref{fig:randominit}. Each bar graph in Figure~\ref{fig:randominit} is an average accuracy score of three GNN models (GCN, GraphSAGE, and GAT) on the respective datasets. 
Our results demonstrate that the NLP embeddings significantly increase the model's performance (up to 3$\times$) because the GNN model learns both the structural information and the node's properties. With the random embeddings model can only learn from the graph structure and cannot perform well in the multi-class classification task. 

\begin{table}[t]
\centering
\caption{Selected Sentence Transformer Models}
\small
\begin{tabular}{lrrr} 
\toprule
Dataset & Emb dim & Avg. Acc & Model size \\ 
\midrule
\texttt{all-MiniLM-L6-v2} & $384$ & $58.80\%$ & $80$ MB \\
\texttt{distiluse-base-multilingual} & $512$ & $45.59\%$ & $480$ MB \\
\texttt{all-mpnet-base-v2} & $768$ & $63.30\%$ & $420$ MB \\
\texttt{all-roberta-large-v1} & $1,024$ & $61.64\%$ & $1,360$ MB \\
\bottomrule
\end{tabular}
\label{tab:nlpmodels}
\end{table}

\begin{figure}[t]
    \centering
    \includegraphics[width=8cm]{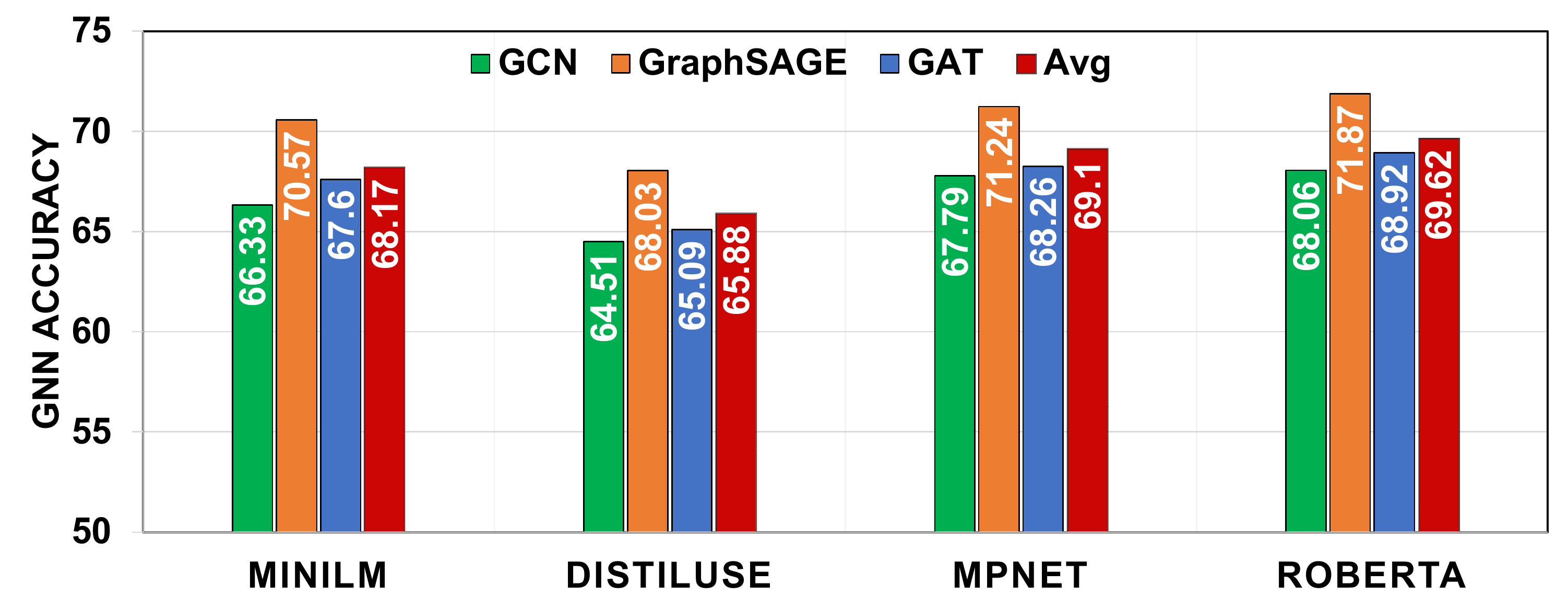}
    \caption{GNN accuracy on \texttt{IGB-HOM-tiny} using different NLP embedding models for initializing node embeddings.}
    \label{fig:embeddmodel}
    \vspace{-3ex}
\end{figure}

\textbf{Selecting Right Language Model For Embeddings}
The quality of embeddings generated by the language model can have a significant effect on the GNN model accuracy. 
To this end, we evaluated the GNN accuracy using a number of widely used language models (see Table~\ref{tab:nlpmodels}) using the sentence transformer package provided by the Huggingface library~\cite{huggingface}. 
Table~\ref{tab:nlpmodels} summarizes the average language model performance on NLP tasks, their respective model size, and generated text embedding dimensions.  

Using these language models, we evaluate the performance of GNN models on the IGB-tiny dataset. To do this, we generated embeddings using each of the language models independently and trained all the GNN models using the respective language embeddings. 
Figure~\ref{fig:embeddmodel} shows the impact of these language models on the GNN accuracy. The average accuracy is calculated by averaging the reported accuracy of GCN, GraphSAGE, and GAT models. 
Although with this evaluation, it is not possible to concisely determine if the large performance range is due to the accuracy of the model or the embedding dimensions, it is clear that the language model itself has a direct impact on the GNN performance. 
In our next evaluation, we will isolate the impact of the model by reducing the dimensions of the embeddings. 

\begin{figure}[t]
    \centering
    \includegraphics[width=8cm]{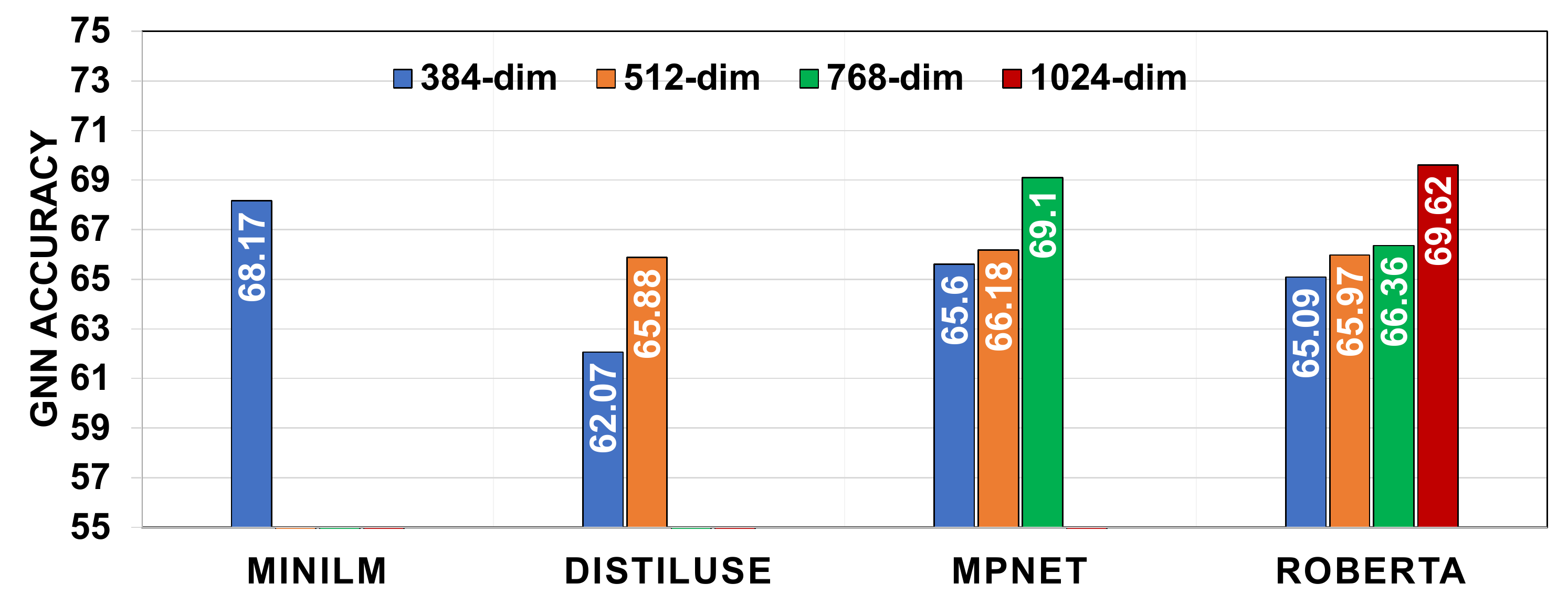}
    \vspace{-2ex}
    \caption{Average GNN accuracy on \texttt{IGB-HOM-tiny} dataset using normalized embeddings across different language models. The average accuracy is computed as across the GCN, GraphSAGE, and GAT models.}
    \label{fig:dimstudy}
\end{figure}

\begin{figure}[t]
    \centering
    \includegraphics[width=8cm]{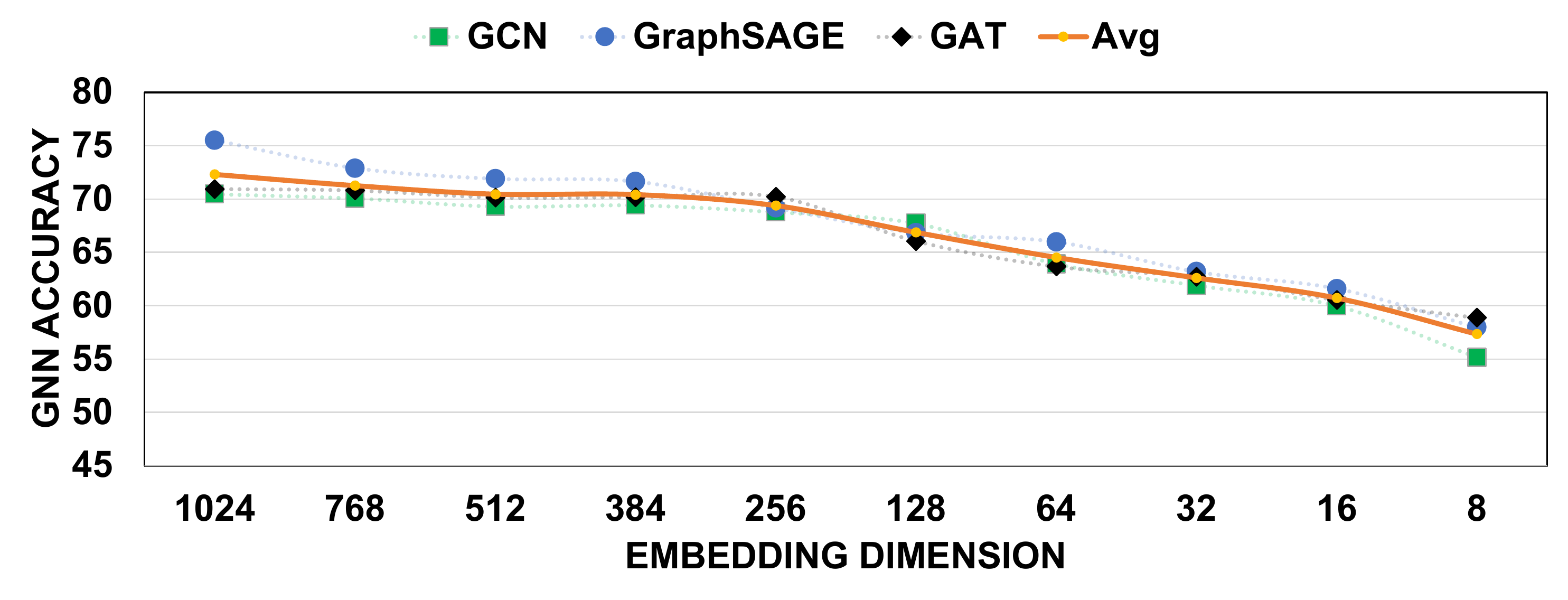}
    \vspace{-2ex}
    \caption{GNN accuracy on RoBERTa~\cite{roberta} embeddings when the dimensions are reduced from 1024 to 8 for \texttt{IGB-HOM-small}. Higher embedding dimensions provide better performance.}
        \label{fig:extremedimstudy}
        \vspace{-3ex}
\end{figure}

\textbf{Impact Of Embedding Dimension}
We normalized the different-sized embeddings from various language models using principal component analysis (PCA), a dimensionality reduction technique~\cite{pca}. 
We used the GPU implementation of PCA from the NVIDIA cuML library~\cite{cuml}. 
The results of the experiments using GCN, GraphSAGE, and GAT models with varying embedding dimensions are presented in Figure~\ref{fig:dimstudy} and Figure~\ref{fig:extremedimstudy}. 

From Figure~\ref{fig:dimstudy}, both mpnet~\cite{mpnet} and RoBERTa~\cite{roberta} models outperform the other language models while training GNN. This is because both these NLP models are inherently better language models (see Table~\ref{tab:nlpmodels}) and their respective accuracy improvements are reflected in the GNN model's accuracy. 

From the figures, reducing the embedding dimensions using PCA had an impact on the accuracy across all the GNN models. 
This is not surprising as the PCA is a lossy pruning technique. 
For RoBERTa~\cite{roberta} embeddings, reducing the dimensions from 1024 to 384 results in a 3.55\%, 1.47\%, and 0.58\% reduction in accuracy for the GCN, GraphSAGE, and GAT models, respectively. 
The GAT model was found to be more resilient to reductions in dimensionality when compared to the GraphSAGE and GCN models. 
A similar observation is noted in other language models. 

Reducing the dimensions from 1024 to 384 results in at least 2.67$\times$ memory capacity savings during training and inference operation with up to 3.55\% loss in accuracy for RoBERTa embeddings. 
Further reducing the dimensionality offers significant memory footprint reduction but results in a significant drop in the GNN model performance as shown in Figure~\ref{fig:extremedimstudy}. 
A similar set of observations are also seen with different IGB datasets including \texttt{IGB-HET} graphs and are not reported. 
Based on this analysis, the IGB dataset collection provides downloadable embeddings from the RoBERTa language model and provides a toolkit to generate other embeddings in our open-source codebase~\cite{dummygithub}.

\section{GNN Input Language Influence}
\label{sec:gnnlanguage}
Scientific databases such as MAG~\cite{mag} and SemanticScholar~\cite{semanticscholar} consist of papers that are written in more than 80 languages. 
For instance, MAG~\cite{mag} database has more than 12 languages with over 1 million papers and only $~50$\% of the total papers are written in English. 
In order to understand the impact of language on the accuracy of GNN models, it is important to consider the language models used to generate node embeddings, which are typically trained on web corpus data~\cite{roberta, gpt2}.

Our goal is to determine if the language used in the dataset to train the NLP model makes any difference in the GNN performance.
To do this, the study would require a language model trained exclusively on a specific language. 
However, finding such a pre-trained large language model is currently impossible. 
As a result, the study used multilingual language models that included or excluded specific languages.

We generated two types of embeddings: one using a language model trained in English and the other using a multi-lingual language model that included Japanese, Spanish, and French (multi-lingual embedding). 
The results, shown in Table~\ref{tab:langperf}, indicate that there is no significant performance improvement achieved by including specific languages, regardless of the model dimensions and language model used. 
We found similar results when running different embedding dimension models on the French and Spanish datasets. 
This makes us believe that the GNNs models are likely language agnostic.

\begin{table}[h]
\centering
\caption{Average GNN accuracy with various languages. Language models have little effect on the accuracy of GNNs on different language inputs.}
\begin{tabular}{lrrr} 
\toprule
Model* & \texttt{IGB-japanese} & \texttt{IGB-spanish} &  \texttt{IGB-french} \\ 
\midrule
$384-eng$ & $62.89$ & $59.77$ & $60.39$ \\
$384-all$ & $62.83$ & $59.01$ & $59.48$ \\
\hline
Average & \boldmath{$62.86_{\pm0.03}$} &  \boldmath{$59.39_{\pm0.38}$} &  \boldmath{$59.94_{\pm0.46}$} \\
\hline
$768-eng$ & $64.82$ & $61.63$ & $61.15$  \\
$768-all$ & $64.74$ & $61.77$ & $61.25$  \\
\hline
Average & \boldmath{$64.78_{\pm0.04}$} &  \boldmath{$61.70_{\pm0.07}$} &  \boldmath{$61.20_{\pm0.05}$} \\
\hline
$512-v1$ & $61.82$ & $58.86$ & $57.48$  \\
$512-v2$ & $61.61$ & $58.42$ & $57.04$  \\
\hline
Average & \boldmath{$61.72_{\pm0.11}$} &  \boldmath{$58.64_{\pm0.22}$} &  \boldmath{$57.26_{\pm0.22}$} \\
\bottomrule
\end{tabular}
\label{tab:langperf}
\end{table}

\subsection{Overall Summary}
The overall results of the \pname{} datasets for all the GNN models with two complex tasks are summarized in Table~\ref{tab:igbhomogeneousresult} and Table~\ref{tab:igbheterogeneousresult}.
The \texttt{IGB-HOM-tiny} and \texttt{IGB-HOM-small} datasets are trained for 20 epochs while the rest are trained for three epochs due to the long training time. 
The accuracy of the models generally decreases in the more complex 2983 tasks as the model must identify finer differences in the nodes to classify them into a large number of classes. 
We observe up to 14.93\% drop in performance for \texttt{IGB-HOM} dataset. 
Longer training (more epochs) may help recover some of these accuracy drops, but ideally, better models are needed to handle the finer classification task. 

Surprisingly the relational GNN models are tolerant to complex multi-class problems on the \texttt{IGB-HET} dataset. 
Despite an increase in task complexity from 19 classes to 2983 classes, the accuracy of relational GNN models does not decline. 
We believe the extra structural information present in the heterogeneous graph enables these models to classify with greater precision. 
Further investigation is necessary to fully comprehend this phenomenon. 

\subsection{Existing System Limitations}
The existing system is unable to handle the complexity of training GNN models using the full IGB dataset. 
This is due to the large size of the embedding tables and graphs (see Table~\ref{tab:igbhomogeneous} and Table~\ref{tab:igbheterogenos}), which require large memory capacity. 
Despite the use of state-of-the-art systems and software, training GNN models on the full IGB dataset on a single system is still a challenge and time-consuming task.
Let's briefly describe how GNN training occurs before discussing the system limitations.  

GNN training operation can be split into three key stages: sampling, aggregation, and computation. 
The first stage in GNN training is to sample nodes or sub-graphs from the graph to form a mini-batch. 
In the aggregation stage, information from the neighborhood of each node present in the mini-batch is aggregated to form a node representation. 
This process involves reading node embeddings and forming an aggregated representation for the node.
In the final stage, the aggregated node representation is fed into a neural network to train the model. 

As graphs and embeddings come in varying sizes, frameworks such as DGL~\cite{dgl} and PyG~\cite{pyg} provide a different mechanism to optimally place graphs and embeddings in the system for efficient training execution.
If the dataset fits in the GPU memory (e.g. \texttt{IGB-HET-tiny} and \texttt{IGB-HET-small}), then both graphs and embeddings can be preloaded to the GPU memory and then GNN models can be trained. 
Our observation is that, when the datasets fit in the GPU memory, we achieve up to 80\% GPU utilization (average 50\%) during training. 

If the dataset fits in the host CPU memory(e.g. \texttt{IGB-HET-medium} and \texttt{IGB-HET-large}), the graph is either placed in the GPU memory or the host memory, depending on the size of the graph, while the embeddings are loaded in the host memory. 
During the training operation, the GPU threads can directly sample the graphs from GPU memory and issue zero-copy memory-mapped I/O access to the embeddings using the DGL-UVA~\cite{emogi, pytorchdirect} technique for efficient execution. 
Our measurement shows that, in this case, we achieve up to 80\% GPU utilization (average 50\%). 

\begin{table}
\centering
\caption{\texttt{IGB-HOM} datasets benchmark results using the same GNN model parameters (*trained for 3 epochs). \texttt{N.G.} stands for cannot finish training within time limit.}
\begin{tabular}{lcccccc} 
\toprule
Model & \multicolumn{2}{c}{GCN} & \multicolumn{2}{c}{SAGE} & \multicolumn{2}{c}{GAT}\\
Dataset (\# class) & 19 & 2983 &  19 & 2983 & 19 & 2983  \\ 
\midrule
\texttt{IGB-HOM-tiny}       & $68.06$ & $53.13$ & $71.87$ & $59.49$ & $68.92$ & $51.74$\\
\texttt{IGB-HOM-small}      & $70.46$ & $63.28$ & $75.49$ & $68.70$ & $70.93$ & $63.70$ \\
\texttt{IGB-HOM-medium*}    & $70.63$ & $62.70$ & $70.35$ & $62.55$ & $70.06$ & $61.81$ \\
\texttt{IGB-HOM-large*}     & $50.29$ & $N.G.$  & $64.89$ & $N.G.$ & $64.59$  & $N.G.$ \\
\texttt{IGB-HOM*}      & $48.59$ & $N.G.$  & $54.95$ & $N.G.$ & $55.51$  & $N.G.$ \\
\bottomrule
\end{tabular}
\label{tab:igbhomogeneousresult}
\end{table}

\begin{table}
\centering
\caption{\texttt{IGB-HET} datasets benchmark results using the same GNN model parameters(* trained for 3 epochs). \texttt{N.G.} stands for cannot finish training within time limit.}
\begin{tabular}{lcccccc} 
\toprule
Model & \multicolumn{2}{c}{RGCN} & \multicolumn{2}{c}{RSAGE} & \multicolumn{2}{c}{RGAT}\\
Dataset (\# class) & 19 & 2983 &  19 & 2983 & 19 & 2983 \\ 
\midrule
\texttt{IGB-HET-tiny}      & $66.73$ & $67.66$ & $67.79$ & $68.84$ & $66.80$ & $67.77$\\
\texttt{IGB-HET-small}     & $71.35$ & $71.64$ & $72.54$ & $72.60$ & $72.61$ & $72.27$\\
\texttt{IGB-HET-medium*}   & $71.85$ & $71.79$ & $72.65$ & $72.86$ & $72.56$ & $72.74$\\
\texttt{IGB-HET-large*}    & $N.G.$  & $N.G.$  & $N.G.$  & $N.G.$  & $N.G.$  & $N.G.$\\
\texttt{IGB-HET*}     & $N.G.$  & $N.G.$  & $N.G.$  & $N.G.$  & $N.G.$  & $N.G.$\\
\bottomrule
\end{tabular}
\label{tab:igbheterogeneousresult}
\end{table}

\begin{figure}[t]
    \centering
    \includegraphics[width=8cm]{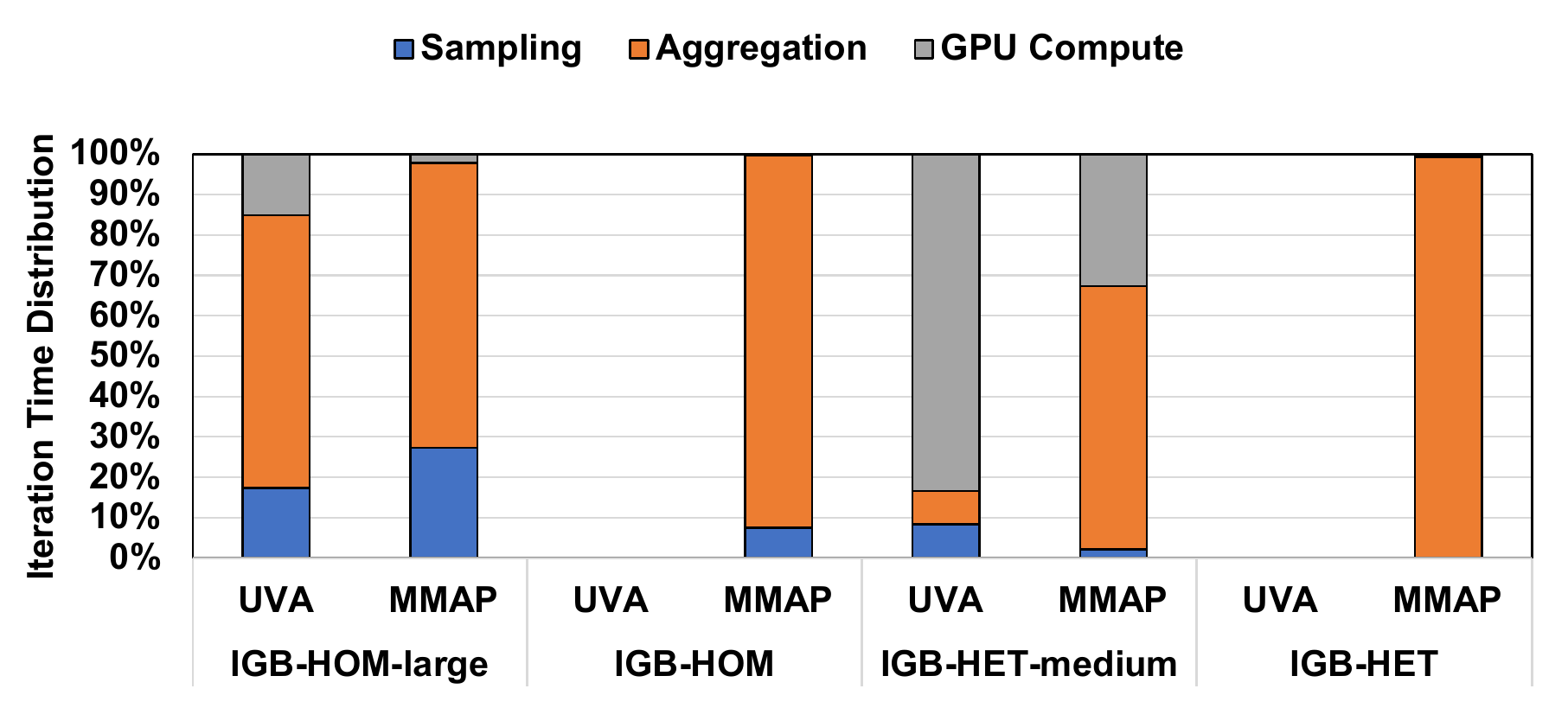}
    \caption{Execution time breakdown of the three stages involved while training GCN and RGCN model. The majority of the time is spent either on the sampling operation or on the aggregation step.}
        \label{fig:systemperf}
\end{figure}

If the graph and embeddings do not fit in the host memory of a single system, as is the case of the \texttt{IGB} dataset, then embeddings can be memory-mapped (\texttt{mmap}) and be stored in fast storage medium like NVMe SSDs.  
The intuition here is that, even though each mini-batch training requires a small working set size ($\sim$600MB for \texttt{IGB-HET} in the GPU memory, the entire graph and embeddings need to be accessible during a training operation. 
This allows the frameworks like DGL~\cite{dgl} and PyG~\cite{pyg} to work on the embedding tables that exceed the host memory capacity of a single system. 

Figure~\ref{fig:systemperf} illustrates the distribution of the time spent for each of the three stages of training GCN and RGCN models on various IGBs graphs. As shown in Figure~\ref{fig:systemperf}, the use of \texttt{mmap} approach leads to a considerable slowdown, even when the graphs and embeddings fit in the host memory, as much as 4.5$\times$ (see IGB-large column). 
Unlike UVA, the \texttt{mmap} abstraction requires the CPU threads to perform the sampling and aggregation stages, accessing the embeddings stored in NVMe SSDs through the operating system page cache, leading to significant overhead. 
For \texttt{IGB-HET} graphs, the node aggregation stage consumes the most significant fraction of iteration time when using the \texttt{mmap} approach, causing low average GPU utilization, less than 5\%. 
Profiling reveals that the \texttt{mmap} approach can only achieve up to 25\% of storage bandwidth (1GBps) and is mainly limited by the system's page fault handler and page-cache throughput.


\section{Discussion}

The main motivation for IGB datasets creation is to have a dataset that can be used by academia and the industry for system designs and GNN model development. ~\pname{} enables:

\textit{Better experiments for GNN model development:}
IGB is created to highlight the inefficiency in existing GNN models. As shown in $\S$\ref{sec:result}, most widely used GNN models only provide a marginal benefit over the naive MLP baseline when datasets are fully labeled, indicating that much more research is needed to further improve GNN model's accuracy.

\textit{Raw text and embedding:}
IGB addresses the lack of large-scale graph datasets with rich labeling information and adaptable node features. This dataset also offers GNN model developers access to raw text data which is crucial for training emerging GNN models like GraphFormer~\cite{graphformers} and GLEM~\cite{glem}. This also enables evaluating the importance of embedding generation and the significance of embedding quality during GNN model training. The IGB dataset is currently the only large-scale dataset for these purposes.

\textit{System scaling:}
IGB is designed to cater support for emergent research and engineering topics for applying GNN in the industry as well. Most of the existing datasets focus on GNN model performance while IGB, in-additional to model performance, enables scalable and efficient system and framework design for GNN deployment in the industry. The IGB presents new system design challenges that previous datasets could not capture. For instance, when training with the DGL on the SAGE model using the OGB MAG240M dataset, we never encountered out-of-memory (OOM) issues as the dataset fits within the $<<1$TB CPU memory (the largest single GPU instance in \texttt{AWS - p4dn.24x} has only $1$TB memory).

However, training with the full IGB datasets, whether homogenous or heterogeneous, on a single CPU server is impractical and requires the development of efficient distributed GNN training methodologies or efficient access to storage. This necessitates the discovery of efficient graph partitioning schemes, communication primitives, and evaluating systems with strong and weak scaling studies. Many of these experiments cannot be effectively conducted due to the lack of large, scalable, and meaningful datasets. The IGB dataset collection addresses these requirements head-on.

\section{Conclusion}

This work introduces \pname{}, a research tool for deep learning practitioners to thoroughly evaluate and test GNN models with accuracy.
It provides access to a large graph dataset that includes both homogeneous and heterogeneous graphs, with more than 40\% of their nodes labeled. 
\pname{} has been designed to be flexible, offering options for the examination of various GNN architectures, embedding generation techniques, and analyzing system performance while training or inferencing GNN models.
IGB is open-sourced, compatible with DGL and PyG frameworks, and includes raw text data that can inspire new research at the intersection of natural language processing and graph neural network research topics. 

\section{ACKNOWLEDGMENTS}

We would like to acknowledge all of the help from members of {the} IMPACT research group, NVIDIA GNN team, and AWS Graph ML team without which we could not have achieved any of the above reported results. 
Special thanks to Dr. Zaid Qureshi from NVIDIA Research and Jeongmin Park from the IMPACT research group for their valuable suggestion on writing and performance evaluations. 
We also would like to acknowledge feed-backs received anonymous reviewers and Dr. Piotr Bigaj from NVIDIA. 
This work is partly funded by the Amazon Research Awards. 
This work uses GPUs donated by NVIDIA, and is partly supported by the IBM-ILLINOIS Center for Cognitive Computing Systems Research (C3SR) and the IBM-ILLINOIS Discovery Accelerator Institute (IIDA).
The datasets in this work is released to public in cooperation with Amazon using AWS Open Data Sponsorship Program.

\bibliographystyle{ACM-Reference-Format}
\bibliography{main}

\end{document}